\pdfoutput=1

\documentclass[11pt]{article}

\usepackage[final]{acl}
\usepackage{times}
\usepackage{latexsym}
\usepackage{appendix}
\usepackage{bbm}
\usepackage{lipsum}
\usepackage{amsmath}

\usepackage{ulem}
\usepackage{times}
\usepackage{latexsym}

\usepackage[T1]{fontenc}

\usepackage[utf8]{inputenc}

\usepackage{microtype}
\usepackage{graphicx,amsmath,subfigure}
\usepackage{todonotes}
\usepackage{caption}
\usepackage{geometry}
\usepackage{float}
\restylefloat{table}
\usepackage{ulem}
\usepackage{caption}
\usepackage{booktabs}
\usepackage{arydshln}
\usepackage{comment}
\title{CILDA: Contrastive Data Augmentation using Intermediate Layer Knowledge Distillation}

\author{Md Akmal Haidar$^1$\thanks{\hspace{3mm}Work done while at Huawei.} \hspace{3mm} Mehdi Rezagholizadeh$^1$ \hspace{3mm} Abbas Ghaddar$^1$\hspace{3mm} Khalil Bibi$^1$ \\\textbf{Philippe Langlais$^2$ \hspace{3mm} Pascal Poupart$^3$}\\
$^1$Huawei Noah’s Ark Lab\\
$^2$ RALI/DIRO, Universit\'e de Montr\'eal, Canada\\
$^3$ David R. Cheriton School of Computer Science, University of Waterloo \\
\texttt{\{mehdi.rezagholizadeh,abbas.ghaddar\}@huawei.com}\\ \texttt{felipe@iro.umontreal.ca, ppoupart@uwaterloo.ca} }

\newcommand{\bert}{\textsc{Bert}}

\begin{document}
\maketitle

\begin{abstract}

Knowledge distillation (KD) is an efficient framework for compressing large-scale pre-trained language models. Recent years have seen a surge of research aiming to improve KD  by leveraging \textbf{C}ontrastive Learning, \textbf{I}ntermediate \textbf{L}ayer Distillation, \textbf{D}ata Augmentation, and \textbf{A}dversarial Training. 
In this work, we propose a learning based data augmentation technique tailored for knowledge distillation, called CILDA. To the best of our knowledge, this is the first time that intermediate layer representations of the main task are used in improving the quality of augmented samples. More precisely, we introduce an augmentation technique for KD based on intermediate layer matching using contrastive loss to improve masked adversarial data augmentation. CILDA outperforms existing state-of-the-art KD approaches on the GLUE benchmark, as well as in an out-of-domain evaluation. 

\end{abstract}

\section{Introduction}

The exponentially increasing size of pre-trained large language models~\cite{Bert,Roberta,raffel2020exploring,brown2020language} has been a persistent concern regarding the efficiency and scalability of Natural Language Understanding (NLU) in real world applications. Knowledge Distillation (KD)~\cite{bucilua2006model,KD} is a technique for transferring the knowledge from a large-scale model (called teacher) to a smaller one (called student), so that the latter model can be employed on edge device~\cite{Distilbert,tang2019distilling,mukherjee2020xtremedistil,li2021select}. This is done by minimizing the KL divergence between the teacher and student probabilistic outputs. 

Numerous techniques have been exploited recently to increase the knowledge transfer beyond logits matching. For instance, it has been found beneficial to perform distillation on the internal components (parameters) of the teacher and student, which is known as Intermediate Layer Distillation~\cite{PKD,Mobile,ALP-KD,wang2020minilmv2,wang2020minilm,LRC, wu2021universal}.

Data Augmentation has also been successful for  KD~\cite{Tiny,Cutoff,CODA}, as researchers have found that the student has less opportunity to acquire useful information from the teacher when limited data are available for training~\cite{kamalloo2021not,kamalloo2022chosen,jafari2021knowledge}. Adversarial Training was also employed in KD~\cite{FreeLB,rashid2020towards,MATE,he2021generate} to improve the robustness and generalization, as the student may predict inconsistent outputs with slight distortion to the data distributions~\cite{li2021select}. Recently, Contrastive Learning~\cite{gutmann2010noise,hjelm2018learning,arora2019theoretical} has been exploited for improving knowledge transfer~\citep{CRD}, and to optimize the intermediate layer mapping scheme~\cite{CODIR}.  

Each of the aforementioned training techniques has proven to be effective in addressing a specific challenge in KD. However, we are not aware of a single method that takes advantage of all of them. In this paper, we propose CILDA, a KD method that incorporate \textbf{C}ontrasting Learning, \textbf{I}ntermediate \textbf{L}ayer Distillation, \textbf{D}ata Augmentation, and \textbf{A}dversarial Training. A contrastive loss is employed to improve the information flow in intermediate layer distillation, as well as the generation of adversarial examples for KD training. CILDA delivers new state-of-the-art results on 6-layer \bert{} model compression, on the GLUE benchmark~\citep{GLUE}, as well as outperforming other KD methods in out-of-domain evaluation.

\section{Related Work}

Many studies~\citep{jawahar2019does,BertP,BertD} have noticed that important structural linguistic information are hidden in the intermediate layers of Transformer models~\cite{vaswani2017attention}. Recent KD methods propose to match teacher and student: intermediate layers representations~\cite{Tiny,PKD,Mobile,wu2020skip}, embedding matrix~\cite{Distilbert}, and self-attention distributions~\cite{wang2020minilmv2,wang2020minilm}.  Other variants of KD methods have been proposed such as Annealing-KD~\cite{jafari2021annealing} and Pro-KD~\cite{rezagholizadeh2021pro}, two stage distillation methods where a smooth and gradual training of the student is controlled by a dynamic temperature factor, followed by a simple cross entropy loss for a few epochs.

Augmented adversarial examples~\cite{miyato2016adversarial} are label-preserving transformations in the embedding space that are used to improve generalizability of models. FreeLB~\citep{FreeLB} is an adversarial algorithm which creates virtual adversarial examples from  word embeddings, and then performs the parameter updates on these adversarial embeddings. MATE-KD~\cite{MATE} is a min-max adversarial data augmentation approach for KD, where an extra \textit{generator} model is trained to generate adversarial text by maximizing the logit output margins between the teacher and the student models.

Contrastive learning is a self-supervised representation learning method ~\citep{SimCLR,CODA,Oord} which learns the feature representation of the samples by contrasting positive and negative samples. CODIR~\cite{CODIR} is a contrast-enhanced diversity promoting method between teacher and student intermediate representations of data samples from the same class. MATE-KD is the most related work to our solution, with one notable difference: to the best of our knowledge, our technique is the first to deploy intermediate layers with the contrastive objective in the data augmentation process.

\section{CILDA} 
\begin{figure}[!htp]
    \centering
    \includegraphics[scale=0.31]{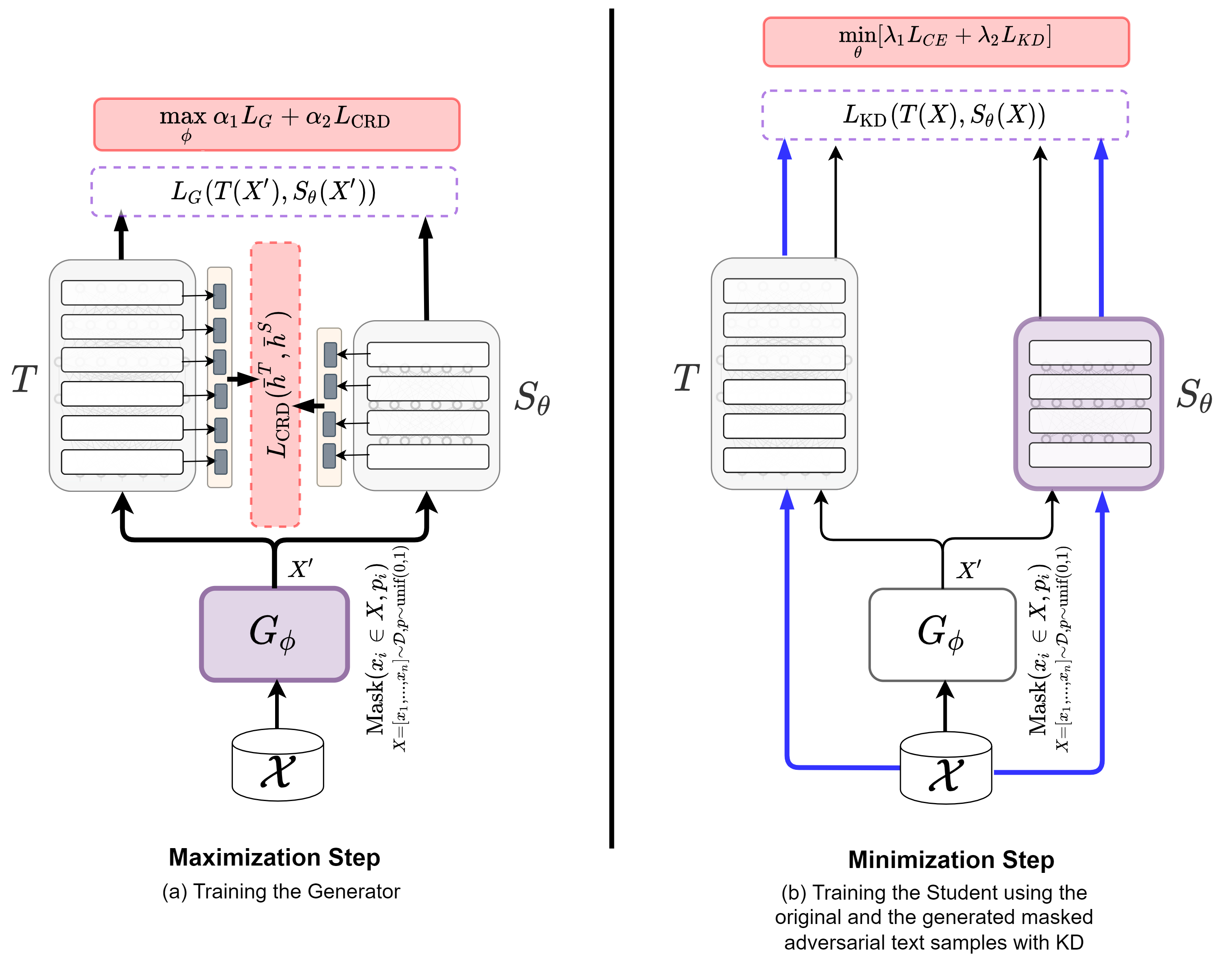}
    \caption{Illustration of maximization and minimization steps of CILDA.}
   \label{fig:plot}
\end{figure}

In this section, we introduce CILDA, our contrastive approach for masked adversarial text augmentation for knowledge distillation using intermediate layer matching. Inspired by~\citep{MATE}, we deploy a generator (e.g. BERT) which will be trained to map masked inputs, $\tilde{X}$, to augmented samples, $X'$. The objective of this mapping is to perturb the inputs (in their vicinity) such that their corresponding output and intermediate layer representations of the teacher and student networks diverge to their maximum. Generating such maximum divergence augmented samples aims to fill the existing major gaps in the training data. 
Inspired by MATE-KD~\citep{MATE}, we mask input tokens with a certain pre-defined probability, $p$. 
The architecture of our model is depicted in Figure~\ref{fig:plot}. Our training is comprised of two alternating steps we describe hereafter.

\paragraph{Maximization Step: Generating Augmented Samples} In the maximization step, the generator is trained in a way that the difference between the teacher and the student are maximized. In opposed to the MATE-KD technique which only evaluates the divergence of the student and teacher networks based on their output, our technique takes intermediate layer matching into account as well. To the best of our knowledge this is the first time that the distance of intermediate layer representations are considered in the data-augmentation generation process. To be concise, MATE-KD only pays attention to the distance of samples in the output space, while our technique concerns the distance of samples in the input space as well. We hypothesize that to identify maximum divergence augmented samples, both input feature distances and output predictions are important. Our CILDA loss function to train the generator can be described as:
\begin{equation}
\begin{split}
& L_{G_\phi}=\alpha_1 L_G+\alpha_2 L_{CRD} \\
& L_G=KL \Big( \sigma(\frac{T(X')}{\tau_1}),\sigma(\frac{S_\theta(X')}{\tau_1}) \Big)
\end{split}
\label{eq:kl}
\end{equation}
where $L_G$ is the KL-divergence loss between the teacher and the student logits, $T$ and $S_\theta$ represent the teacher model and the student model with $\theta$ parameters respectively, $\sigma$ is the softmax function and $\tau_1$ is the temperature parameter that controls the softness of the output distributions, $\alpha_1$ and $\alpha_2$ are hyper-parameters. $X'$ is the adversarial text output obtained by applying argmax to the generator output in the forward pass. It is worth mentioning that due to the non-differentiability issue of argmax in the backward pass, we use Gumbel-Softmax ~\citep{Gumbel} at the output of the generator. More details can be found in~\citep{MATE}.
$L_{CRD}$ is the contrastive distillation loss that we introduced to the maximization step of MATE-KD. This contrastive loss is obtained by using the intermediate representation outputs of the teacher and the student models: 
\begin{equation}
    L_{CRD} = -\log\frac{exp(<\Bar{h}_k^T,\Bar{h}_k^{S_\theta}>/\tau_2)}{\sum_{j=0}^K exp(<\Bar{h}_k^T,\Bar{h}_j^{S_\theta}>/\tau_2)}
    \label{eq:congcrd}
\end{equation}
where $\tau_2$ is the temperature parameter that controls the concentration level~\citep{CODIR}. $\Bar{h}_k^T$ and $\Bar{h}_k^{S_\theta}$ are the intermediate layer representation of the teacher and student networks respectively. $k$ and $j$ are indices of the samples of a mini-batch. $k$ is the index of positive samples (i.e. the k$^{th}$ sample of the mini-batch is sent to both of the student and teacher networks to obtain their representations) and when $j\neq k$, we get negative samples (i.e. any other sample in the mini-batch excluding the k$^{th}$ sample) in a batch of $K$ samples. 
The goal of this objective function is to map the student representations $\Bar{h}_k^{S_\theta}$ of the positive sample $k$ to $\Bar{h}_k^T$, as well as the negative representations $\{\Bar{h}_j^{S_\theta}\}_{j\neq k}^K$  far apart from $\Bar{h}_k^T$. $<.,.>$ is the cosine similarity between two feature vectors. 

For an arbitrary sample $l$ in a mini-batch, the entire intermediate layer representations of the teacher and the student models (e.g. the $<CLS>$ representation of each layer of the networks) are concatenated to form $\hat{h}_l^T=[\Bar{h}_{1,l}^T,\cdots,\Bar{h}_{n,l}^T]$, $\hat{h}_l^{S_\theta}=[\Bar{h}_{1,l}^{S_\theta},\cdots,\Bar{h}_{m,l}^{S_\theta}]$. Then these concatenated representations are further mapped into the same-size lower-dimensional spaces using linear projections $\Bar{h}_l^T,\Bar{h}_l^{S_\theta} \in R^u$ to calculate the distillation loss $L_{CRD}$. Here, $n$ and $m$ denote the number of intermediate layers of the teacher and the student networks respectively.

\paragraph{Minimization Step: Deploying Augmented Samples}  In the minimization step, the augmented adversarial samples produced by the generator and the training samples are used to minimize the difference between the teacher and the student. 
For this step, in the very general form, one can consider to match the student and teacher networks on their outputs and intermediate layer representations (e.g. using the contrastive loss) and the CE loss to match the output of the student with the labels:
\begin{equation}
L_{S_\theta} = \lambda_1 L_{CE} +\lambda_{2} L_{KD} 
\end{equation}
where, $L_{CE}$ describes the cross-entropy loss between the true label. 

\begin{table*}[!thp]
\small
\centering
\begin{tabular}{lccccccccc}

\toprule
\textbf{Model} & \textbf{CoLA}& \textbf{SST-2}& \textbf{MRPC}& \textbf{STS-B}& \textbf{QQP}& \textbf{MNLI}& \textbf{QNLI}& \textbf{RTE}& \textbf{Avg.}\\

\midrule
\multicolumn{10}{c}{\textsc{Dev}}\\
\midrule

Teacher & 68.1 & 96.4 &91.9 & 92.3 & 91.5 & 90.2 & 94.6 & 86.3 & 88.9\\
\hdashline
Vanilla-KD & 60.9 & 92.5 & 90.2 & 89.0 & 91.6 & 84.1 & 91.3 & 71.1 & 83.8\\
Annealing-KD & 61.7 & 93.1 & 90.6 & 89.0 & 91.5 & 85.3 & 92.5 & 73.6 & 84.7\\
MATE-KD & 65.9 & 94.1 & 91.9 & 90.4 & 91.9 & 85.8 & \bf 94.6 & 75.0 & 86.2\\
\hdashline
CILDA & \textbf{67.1} & \textbf{94.7} & \textbf{92.0} & \textbf{90.5} & \textbf{92.1} & \textbf{86.8} & 92.9 & \textbf{76.2} & \textbf{86.5}\\

\midrule
\multicolumn{10}{c}{\textsc{Test}}\\
\midrule

Teacher         & 68.6 & 97.1 & 93.0 & 92.4 & 90.2 & 90.7 & 95.5 & 87.9 & 89.4 \\

\hdashline
Vanilla-KD      & 54.3 & 93.1 & 86.0 & 85.7 & 89.5 & 83.6 & 90.8 & 74.1 & 82.1\\
Annealing-KD    & 54.0 & 93.6 & 86.0 & 86.8 & 89.7 & 84.4 & 90.8 & 73.7 & 82.4\\
MATE-KD         & 56.0 & \bf 94.9 & 90.2 & 88.0 & 89.7 & 85.2 & 92.1 & 75.0 & 83.9\\ %
\hdashline
CILDA           & \bf 56.2 & \bf 94.9 & \bf 90.5 & \bf 89.0 & \bf 89.9 & \bf 86.1 & \bf 92.5 & \bf 77.0 & \textbf{84.5}\\
\bottomrule

\end{tabular}
\caption{\textsc{Dev} and \textsc{Test} performances on GLUE benchmark when RoBERTa$_{24}$ and DistillRoberta$_6$ are used as backbone for the teacher and student variants respectively. Bold mark describes the best results.}
\label{tab:glue_dev_test}

\end{table*}

\section{Experiments}

\subsection{Datasets and Evaluation}
We experiment on 7 tasks from the GLUE benchmark~\cite{GLUE}: 2 single-sentence (CoLA and SST-2) and 5 sentence-pair (MRPC, RTE, QQP, QNLI, and MNLI) classification tasks. Following prior works, we report Pearson correlation on STS-B, Matthews correlation on CoLA, F1 score on MRPC, and use the accuracy for the remaining tasks. For out-of-domain evaluation, we report the performances on HANS~\cite{HANS}, SciTail~\cite{SciTail}, and IMDB using the models finetuned on MNLI, QQP, and SST-2 respectively.

\subsection{Implementation Details}

We use the 24-layer RoBERTa-large~\citep{Roberta} and the 6-layer DistilRoBERTa~\citep{DistilRoberta} as the backbone for the teacher and the student models respectively. We perform hyperparameter tuning, and select best performing models using early stopping on dev sets. We use a linear transformation to map the intermediate representations into a 128-dimensional space and normalized them before computing the loss $L_{CRD}$. For each batch of data, we train the generator for $n_G$ steps and the student model for $n_S=100$ steps. We use $n_G=20$ for CoLA, MRPC, RTE tasks and $n_G$=10 for rest of the tasks. Following~\cite{MATE}, we set $p_{th}=0.3$, $\alpha_1=1, \alpha_2=1, \tau_1=1.0, \tau_2=2.0$  for all of our experiments. We set $\lambda_1$  and $\lambda_2$ to 1/3 for the original training samples. For the augmented samples, we use  $\lambda_{2}=2/9,\lambda_{3}=1/9$ for all tasks. The learning rate and the batch size are tuned from the set of \{1e-5, 2e-5, 4e-6\} and \{8, 16, 32\} respectively.

\subsection{Results and Analysis}
Table~\ref{tab:glue_dev_test} shows the performances of the teacher, baselines, and our method on the GLUE dev and test sets. We compared CILDA to the Vanilla-KD~\cite{KD} baseline, and against 2 strong recently proposed methods~\footnote{We compare with these models because we have published results on GLUE leaderboard using the same teacher and student backbone models.}: Annealing-KD~\cite{jafari2021annealing} and MATE-KD~\cite{MATE}. We observe that CILDA outperforms these models on all GLUE tasks, except on QNLI dev where MATE-KD performs better and SST-2 test where CILDA is on par with MATE-KD. Overall, CILDA outperforms MATE-KD and Annealing-KD by a margin of 0.6\% and 2.1\% on average test set respectively. 

\begin{figure}[ht]
    \centering
    \includegraphics[scale=0.45]{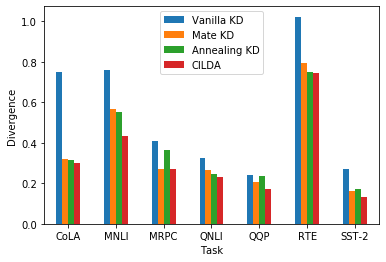}
    \caption{Divergence (lower is better) between the teacher and student logits on GLUE dev sets.}
   \label{fig:divergence}
\end{figure}

We investigate the logits generated by different methods to better understand why CILDA performs better. Figure~\ref{fig:divergence} shows the divergence (lower is better) between the teacher and student logits on GLUE dev sets (except STS-B since it is a regression task) for 4 KD methods. Expectedly, Vanilla-KD (no enhancement) had the maximum divergence with teacher logits (which can be easily distinguished from other methods). We observe that CILDA mimic the teacher better than other methods on all tasks, which may partially explain the performance gains obtained by CILDA. 

\begin{table}[!htb]
\small
\centering
\begin{tabular}{l|ccc}
\toprule
\textbf{Model} & \textbf{HANS} & \textbf{PAWS} & \textbf{IMDB}\\
\midrule
Teacher & 78.2 & 43.3 & 88.9 \\
\midrule
w/o KD & 58.6 & 34.7 & 83.7\\
Vanilla-KD &58.9 & 36.5 & 84.0 \\
Annealing-KD & 61.2 & 35.8 & 84.6 \\
MATE-KD&66.6& 38.3 & 85.0\\
\midrule
CILDA&\textbf{68.1}& \textbf{40.5} & \textbf{85.2}\\
\bottomrule
\end{tabular}

\caption{Out-of-domain performances of models trained on MNLI, QQP, SST-2 and evaluated on HANS, PAWS, and IMDB respectively.}
\label{tab:ood}
\end{table}

Furthermore, we measure the robustness and generalization ability of the tested methods by evaluating them on out-of-domain test sets. Table~\ref{tab:ood} shows performances of models fine-tuned on MNLI, QQP, SST-2 and tested on HANS, PAWS, and IMDB respectively. CILDA significantly outperforms the second best method (MATE-KD) by 1.6\% and 2.2\% on HANS and PAWS respectively, and by a margin of 0.2\% on IMDB.

\section{Conclusion and Future Work}
We proposed a min-max adversarial data augmentation framework for KD, which is powered by contrastive distillation loss for intermediate layer matching. Our algorithm maximizes the intermediate and logit representation margin between the  teacher and the student models. In future works, we would like to investigate the distillation from  super-large models such as Megatron~\cite{shoeybi2019megatron} and T5~\cite{raffel2020exploring}. Also, we would like to improve the generator output quality via distillation from generative models like GPT-2~\cite{GPT}. 

\section*{Acknowledgments}
We thank Mindspore\footnote{\url{https://www.mindspore.cn/}} for the partial support of this work, which is a new deep learning computing framework. 

\normalem

\bibliography{anthology,custom}
\bibliographystyle{acl_natbib}

\end{document}